\UseRawInputEncoding
\documentclass[letterpaper, 10 pt, conference]{ieeeconf}  

\IEEEoverridecommandlockouts                              

\usepackage{longtable}
\usepackage{balance}
\usepackage{booktabs}
\usepackage{algorithm}
\usepackage{bbm}
\usepackage{xcolor}
\usepackage{amsfonts}
\usepackage{amsmath,amssymb,amsfonts}
\usepackage{algorithmic}
\usepackage{graphicx}
\usepackage{textcomp}
\usepackage{xcolor}
\usepackage{makecell}
\usepackage{cite}
\usepackage{graphicx}
\usepackage{amsmath}
\usepackage{amssymb}
\usepackage{booktabs}
\usepackage{multicol}
\usepackage{multirow}
\usepackage{bm}
\usepackage{comment}
\usepackage{booktabs}
\usepackage{array}
\usepackage{hyperref}
\usepackage{cleveref} 
\usepackage{booktabs}
\usepackage{array} 
\usepackage{siunitx} 
\newcommand{\method}{GaussMemory}
\newcommand{\real}{\mathbb{R}}
\newcommand{\vect}[1]{\bm{#1}}
\newcommand{\mat}[1]{\mathbf{#1}}
\newcommand{\loss}{\mathcal{L}}
\newcommand{\set}[1]{\mathcal{#1}}

\definecolor{cmem}{RGB}{52,152,219}
\definecolor{cgauss}{RGB}{231,76,60}
\definecolor{cupdate}{RGB}{46,204,113}
\definecolor{caction}{RGB}{155,89,182}
\definecolor{cllm}{RGB}{243,156,18}
\definecolor{ctemporal}{RGB}{26,188,156}
\definecolor{cpast}{RGB}{149,165,166}
\definecolor{csoft}{RGB}{230,126,34}
\definecolor{cuma}{RGB}{142,68,173}
\Crefformat{figure}{#2Fig.~#1#3}
\Crefmultiformat{figure}{Figs.~#2#1#3}{ and~#2#1#3}{, #2#1#3}{ and~#2#1#3}

\begin{document}
\title{\LARGE \bf
GaussMemory: Task-Driven 3D Gaussian Scene Memory \\for Long-Horizon Robotic Manipulation
{  
}}
\author{Zhiqiang HU, Shouren HUANG and Masatoshi ISHIKAWA
\thanks{}
\thanks{The authors are with the Research Institute for Science \& Technology, Tokyo University of Science
        {\tt\small \{zhiqiang.hu, huang, ishikawa\}@ishikawa-vision.org}}%
}

\maketitle
\thispagestyle{empty}
\pagestyle{empty}

\begin{abstract}
Long-horizon robotic manipulation fundamentally relies on persistent spatial memory. However, existing 3D memory systems function merely as passive recorders: they store observations using fixed, hand-crafted rules, treating every scene element---whether a critical grasp target or an irrelevant background wall---with equal importance. In this paper, we propose a paradigm shift from passive storage to active, task-driven spatial memory. We argue that a robot's memory should not simply record what it sees, but actively learn \emph{how} to remember---discovering which objects to track precisely, how aggressively to update them, and what to discard, all learned end-to-end without hand-designed rules. Crucially, this active paradigm is realized by unifying memory update and readout as two sides of the same cognitive process, enabling bidirectional flow where task needs shape update strategies and vice versa. To instantiate this vision, we introduce \method{}, which leverages 3D Gaussian Splatting as a persistent geometric substrate. On LIBERO, \method{} outperforms MemoryVLA on Goal and Long-10; on VLABench, it surpasses $\pi_0$-FAST by +5.2\% (Track~1) and +6.0\% (Track~6).
\end{abstract}

\section{Introduction}
\label{sec:intro}

\begin{figure}[htbp]
\centering
\includegraphics[width=1\columnwidth]{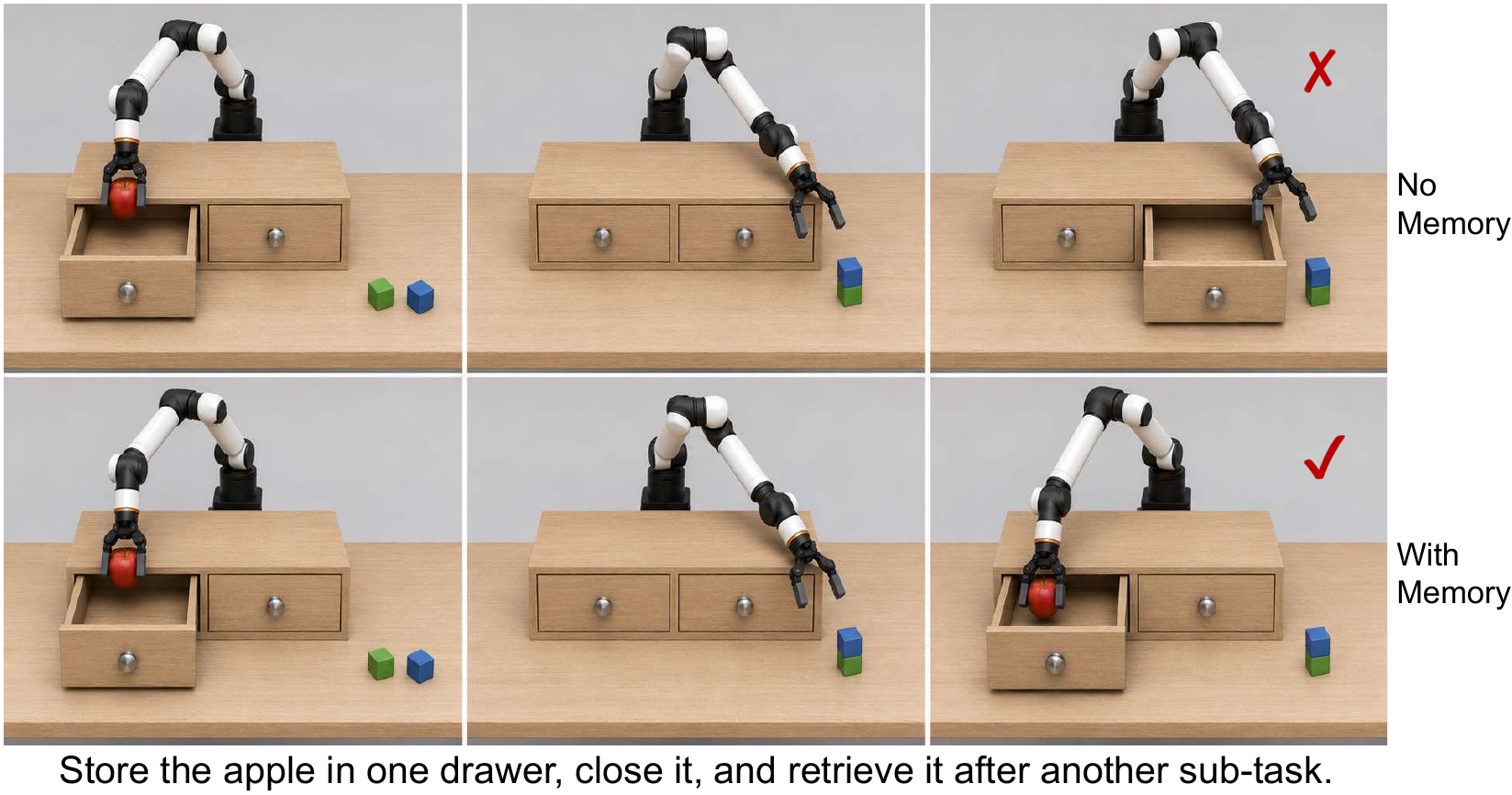}
 
\caption{\textbf{With vs.\ without persistent memory.} Why persistent 3D memory is necessary. The robot stores an apple in one of two visually identical drawers and completes an intervening sub-task. Once both drawers are closed, the current RGB observation no longer reveals the apple's location. A memoryless policy cannot disambiguate the correct drawer.}
\label{fig:conceptb}
\end{figure}

Vision-Language-Action (VLA) models such as OpenVLA~\cite{kim2024openvla} and $\pi_0$~\cite{black2024pi0} have emerged as a powerful paradigm for robotic manipulation, leveraging pretrained vision-language knowledge to map visual observations and language instructions directly to actions. 
However, memoryless VLA systems are mainly limited to tasks in which the next action can be inferred from the current sensory observation. As Fig.~\ref{fig:conceptb} illustrates, when an object is stored in one of two visually identical drawers and subsequently becomes occluded, the current RGB observation is insufficient to identify the correct drawer after an intervening sub-task. Persistent spatial memory is therefore required to retain the object's identity and last observed 3D position across time, enabling reliable retrieval in long-horizon manipulation.

Recent efforts have begun to equip VLA systems with temporal memory, but the solutions remain limited. MemoryVLA~\cite{memoryvla2025} maintains a 2D perceptual-cognitive memory bank achieving strong results, yet its memory is fundamentally 2D and cannot answer spatial queries such as ``where exactly is the plate relative to the glass?'' MEM~\cite{mem2025} introduces multi-scale embodied memory combining semantic, episodic, and procedural stores, but likewise operates in 2D token space without explicit 3D geometry. Language-based approaches such as SayPlan~\cite{rana2023sayplan} record text descriptions of past states but sacrifice geometric precision entirely. Even recent 3D-structured systems such as ConceptGraphs~\cite{gu2024conceptgraphs} that maintain scene graphs with explicit geometry still rely on hand-crafted update rules---thresholded matching, fixed-rate averaging, hard pruning---entirely decoupled from task performance. Across all these categories, the memory systems share a common limitation: they function as \emph{passive recorders} that treat every scene element with equal importance. Crucially, this stems from a deeper issue: memory update and readout are treated as independent operations, with no communication between them.
\begin{figure*}[t]
\centering
\includegraphics[width=0.8\textwidth]{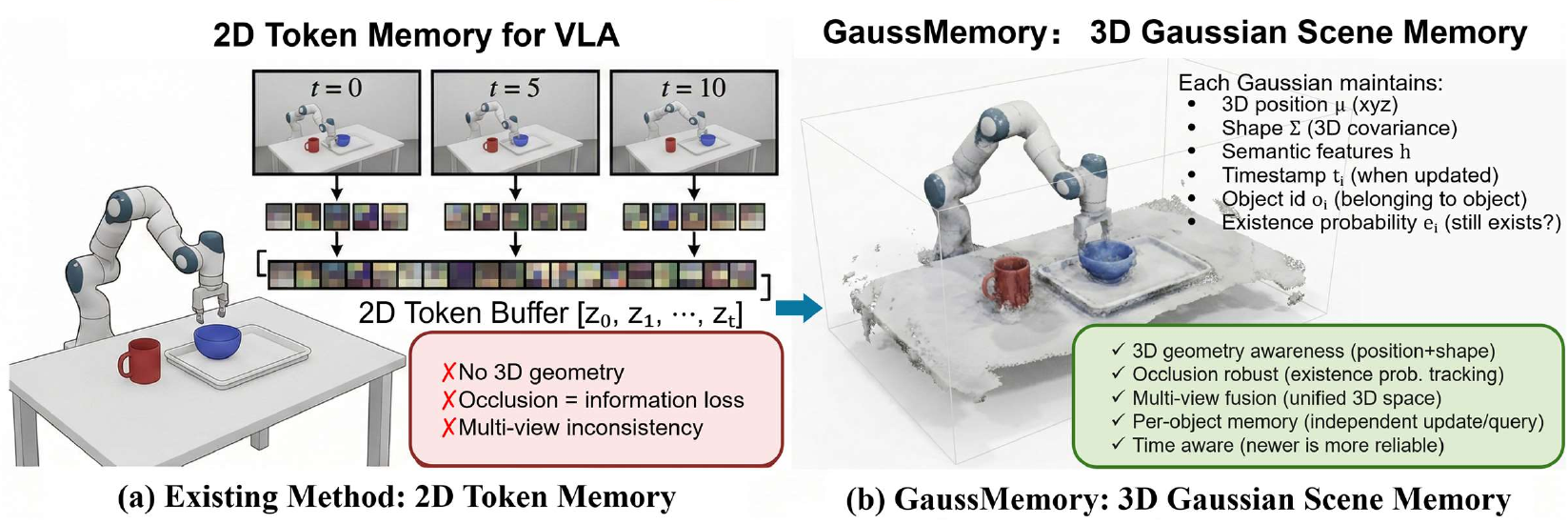}
\caption{\textbf{2D token memory vs.\ 3D Gaussian scene memory.} \textbf{(a)}~Existing VLA methods buffer 2D image tokens that lack 3D geometry, lose information under occlusion, and suffer from multi-view inconsistency. \textbf{(b)}~\method{} maintains a persistent 3D Gaussian Scene Memory where each Gaussian carries explicit position, shape, semantic features, timestamps, object identity, and existence probability.}
\label{fig:concepta}
\end{figure*}
We propose to move beyond this passive paradigm by shifting to \emph{active, task-driven} 3D spatial memory within the VLA framework (Fig.~\ref{fig:concepta}). The memory should learn not just \emph{what} to remember, but \emph{how} to remember---which objects to track with high precision, which observations to trust, when to update aggressively and when to preserve---all guided by one criterion: \emph{does this way of remembering help the robot complete its task?} The key insight is that memory update and readout are not separate processes but two sides of the same cognitive computation: how you \emph{use} memory (readout) should directly shape how you \emph{store} it (update), and vice versa. For instance, if the next sub-task requires placing silverware next to the plate, the readout queries will attend strongly to the plate's position; this attention signal should flow back to bias the update toward tracking the plate more precisely. Conversely, if an object like a cup was just moved, the update event should inform the readout to trust its new position more. This bidirectional coupling---inspired by cognitive science phenomena like the Testing Effect, where retrieval strengthens memory encoding---is absent in existing decoupled systems but enables the task loss to genuinely sculpt memory behavior.

3D Gaussian Splatting (3DGS)~\cite{kerbl20233dgs} is a particularly suitable substrate for realizing active memory within VLA, thanks to three properties: \textit{(1) Natural incrementality}: individual Gaussians can be added, removed, or repositioned without disrupting the rest of the scene. \textit{(2) Explicit geometry}: each Gaussian carries position $\vect{\mu}$, covariance $\mat{\Sigma}$, and opacity $\alpha$, providing the precise spatial information that manipulation planning requires. \textit{(3) Tokenizability}: Gaussian features can be encoded into compact tokens and injected into an LLM backbone alongside language and 2D visual tokens, integrating naturally into the VLA architecture. Yet no existing system takes the step from passive Gaussian storage to active, task-driven Gaussian memory within VLA---where the update mechanism is end-to-end differentiable with respect to the action prediction loss. We realize this unified vision through \textbf{Unified Memory Attention (UMA)}, where observation tokens and readout queries are concatenated and processed in shared cross-attention layers. This design enables bidirectional information flow---the readout is aware of what just changed, and the update is aware of what the task cares about---so that the task's information need shapes memory behavior not only through gradients during training but also through forward attention flow at inference time.

\noindent We present \method{} with three contributions:

\noindent\textbf{Active 3D Gaussian Memory Paradigm}: 
We propose that robot spatial memory should shift from passive recording to active, task-driven learning, instantiated using 3D Gaussian Splatting as a persistent geometric substrate with temporal metadata and an object-centric scene graph. Empirically, the learned position-update coefficient exhibits a bimodal pattern, with an average value of approximately 0.74 for manipulated objects and 0.06 for background Gaussians.

\noindent\textbf{Unified Memory Attention (UMA)}: 
A single cross-attention module that \emph{simultaneously} reads memory for action prediction and writes updates back, making the entire memory lifecycle end-to-end differentiable. UMA enables bidirectional information flow: task needs shape memory updates, and memory state informs action readout.

\noindent\textbf{Persistent 3D Gaussian Memory Integration for VLA}: 
To our knowledge, \method{} is the first to integrate 3D Gaussian Splatting as a \emph{temporal scene memory} within a VLA framework, where persistently updated memory tokens derived from 3D Gaussians via UMA are consumed by an LLM backbone for action prediction. Unlike prior works that encode static Gaussian scenes into tokens for language-based QA~\cite{gaussianvlm2025}, our Gaussian memory is continuously updated across sub-tasks and jointly optimized with robot action loss.

\begin{figure*}[t]
\centering
\includegraphics[width=0.9\textwidth]{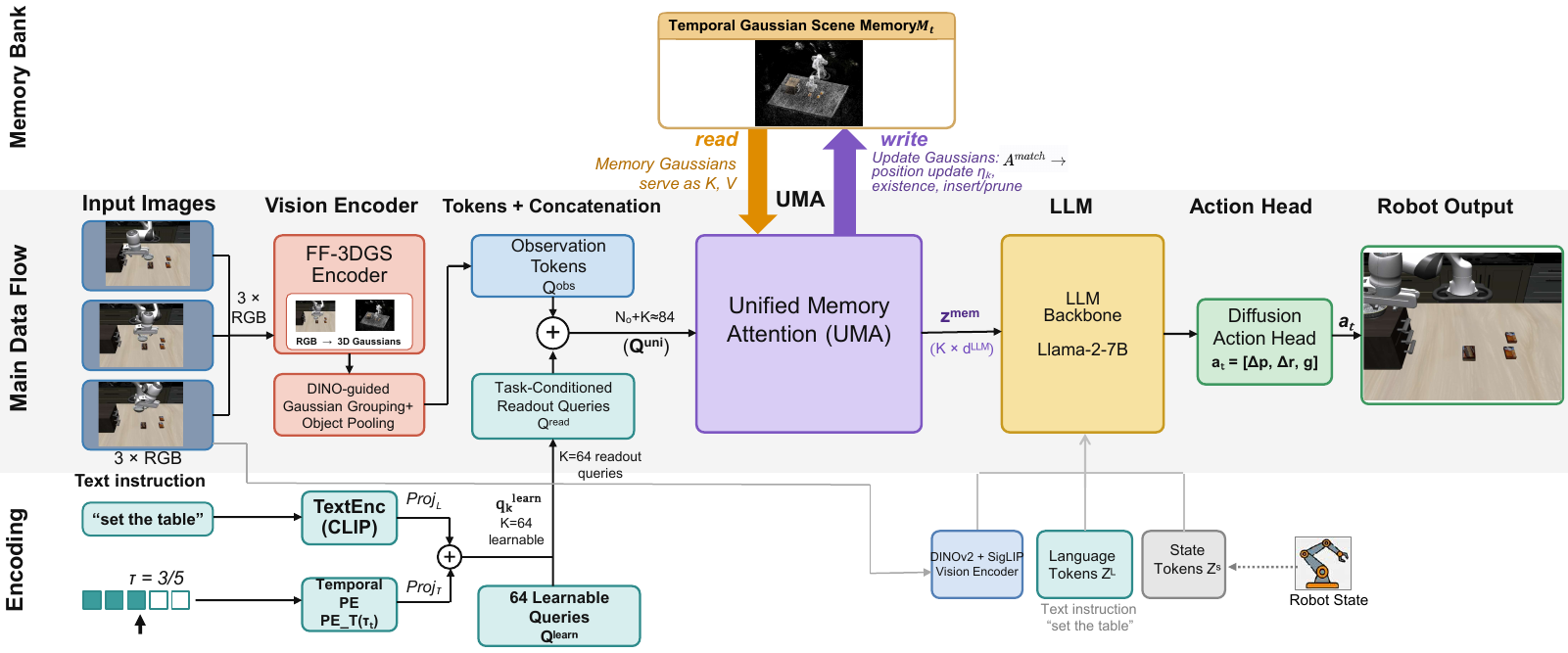}
\caption{\textbf{\method{} architecture.} Multi-view RGB images are encoded by FF-3DGS into observation tokens $\mat{Q}^{\text{obs}}$, which are concatenated with task-conditioned readout queries $\mat{Q}^{\text{read}}$ and jointly processed by \textbf{Unified Memory Attention (UMA)}. UMA cross-attends to the 3D Gaussian Scene Memory (K/V up to 32K persistent Gaussians), producing two parallel outputs: the \emph{output stream} $\mat{Z}^{\text{mem}}$ provides 3D spatial context to the LLM backbone for action prediction via the Diffusion Action Head, while the \emph{update memory stream} writes back updated Gaussians via learned gates $\eta_k$, matching, insertion, and pruning. }
\label{fig:architecture}
\end{figure*}
\section{Related Work}
\label{sec:related}

\subsection{Memory for Long-Horizon Manipulation}

Long-horizon robotic tasks require maintaining information across many interaction steps. Existing memory approaches fall into three categories:

\textit{Language-based memory}: SayPlan~\cite{rana2023sayplan} uses a language-annotated 3D scene graph updated via text descriptions. While compact and LLM-friendly, these approaches lose geometric precision---``the cup is on the counter'' does not encode \emph{where} on the counter or the cup's spatial relationship to nearby objects.

\textit{Image-based memory}: MemoryVLA~\cite{memoryvla2025} maintains a 2D perceptual-cognitive memory bank with temporal positional encoding, achieving strong results on long-horizon tasks. However, its memory is fundamentally 2D token-level---it cannot answer spatial queries such as ``where is the cup relative to the plate?'' that require 3D geometric reasoning.

\textit{3D-structured memory}: ConceptGraphs~\cite{gu2024conceptgraphs} builds object-centric 3D scene graphs from multi-view observations. All these systems separate memory update from memory readout into independent non-differentiable stages.

\method{} differs from all the above by shifting from passive to \emph{active} memory: the update mechanism itself is learned from task outcomes, and 3D Gaussians provide the geometric precision that text and image memories lack.

\subsection{3D Gaussian Splatting for Robotics}

3DGS~\cite{kerbl20233dgs} has expanded into robotic perception.
ManiGaussian~\cite{lu2024manigaussian} uses dynamic Gaussians as a
forward world model, whereas \method{} uses them as a persistent scene
memory for retrospective state tracking and future action prediction.
GaussianVLM~\cite{gaussianvlm2025} encodes static Gaussian scenes for
language-based reasoning and produces text outputs. In contrast,
\method{} maintains a temporally evolving Gaussian memory whose tokens
directly condition robot action prediction.

\section{Method}
\label{sec:method}

\subsection{Problem Formulation}

We consider long-horizon manipulation tasks specified by a high-level language instruction $\set{L}$ that decomposes into a sequence of $T$ sub-tasks $\{\tau_1,\ldots,\tau_T\}$. At each time step $t$, the robot receives multi-view RGB images $\{I_v^t\}_{v=1}^{V}$ and its proprioceptive state $\vect{s}_t$. The goal is to predict actions $\vect{a}_t$ that sequentially complete all sub-tasks while maintaining persistent memory of scene changes.

\subsection{System Overview}

As shown in Fig.~\ref{fig:architecture}, \method{} consists of three components: a \emph{Temporal Gaussian Scene Memory} $\set{M}_t$ (persistent 3D Gaussians), \emph{Unified Memory Attention (UMA)} (simultaneous memory read/write via shared cross-attention), and a Prismatic VLM backbone (Llama-2-7B, shared with OpenVLA/CogACT/MemoryVLA) with a diffusion action head.

\begin{figure*}[t]
\centering
\includegraphics[width=0.92\textwidth]{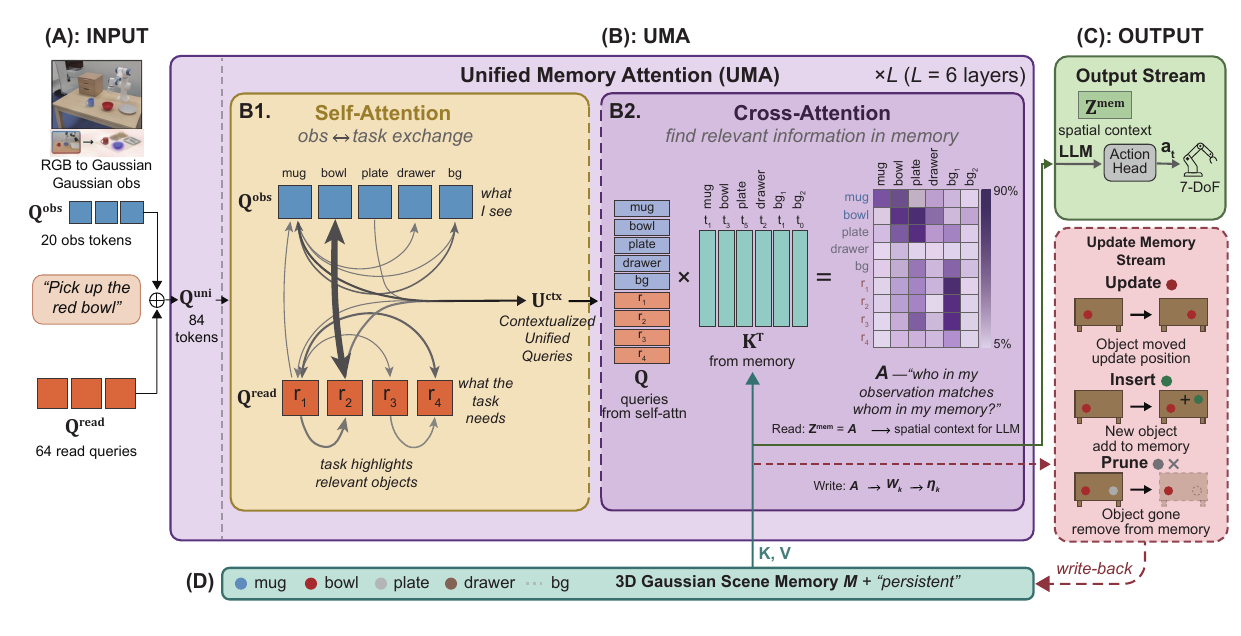}

\caption{\textbf{Inside Unified Memory Attention (UMA),}
expanding the UMA block from Fig.~\ref{fig:architecture}.
\textbf{(A)} Observation tokens and readout queries are concatenated.
\textbf{(B1)} Self-attention exchanges task and perceptual information.
\textbf{(B2)} Shared cross-attention queries Gaussian memory through
keys and values, producing read-query features for
$\mat{Z}^{\mathrm{mem}}$ and observation-to-memory correspondences
$\mat{A}^{\mathrm{match}}$ for memory update.
\textbf{(C)} The two streams support action prediction and memory
update, insertion, and pruning.
\textbf{(D)} Persistent 3D Gaussian Scene Memory.}
\label{fig:dataflow}
\end{figure*}
\subsection{Temporal Gaussian Scene Memory}
\label{sec:memory}

The core data structure is the Temporal Gaussian Scene Memory $\set{M}_t$, a set of augmented Gaussian primitives maintained persistently across the task horizon.

\paragraph{Augmented Gaussian Primitives}
Each Gaussian carries standard 3DGS attributes plus temporal metadata:
\begin{equation}
G_i = (\underbrace{\vect{\mu}_i, \mat{\Sigma}_i, \alpha_i, \vect{c}_i, \vect{h}_i}_{\text{standard 3DGS}},\; \underbrace{t_i^c, t_i^m, \tau_i, o_i,e_i}_{\text{temporal metadata}})
\end{equation}
where $\vect{\mu}_i \!\in\! \real^3$ is position, $\mat{\Sigma}_i$ the covariance, $\alpha_i$ opacity, $\vect{c}_i$ color, $\vect{h}_i$ spherical harmonic features, $t_i^c$/$t_i^m$ creation/modification timestamps, $\tau_i$ sub-task index, and $o_i \!\in\! \{1,\ldots,O\}$ object ID, and existence probability $e_i \in (0,1)$.

\paragraph{Temporal Feature Encoding}
Each Gaussian is encoded as:
\begin{align}
\vect{g}_i^{\text{mem}}
&=
\text{MLP}_{\text{enc}}
\big(
[\vect{\mu}_i;\text{vec}(\mat{\Sigma}_i);
\alpha_i;\vect{h}_i]
\big)
\nonumber\\
&\quad
+\text{PE}_{3D}(\vect{\mu}_i)
+\text{PE}_T^{\text{rel}}(t_i^m,t)
\label{eq:temporal_enc}
\end{align}
where $\text{PE}_{3D}$ is a 3D Fourier positional encoding~\cite{tancik2020fourier}, and $\text{PE}_T^{\text{rel}}(t_i^m,t)=\text{PE}_T(t-t_i^m)$ encodes the relative age of a memory entry.

\begin{equation}
\mathrm{PE}_T(t)
=
\bigoplus_{n=1}^{N_T}
\left[
\sin(\omega_n^T t),
\cos(\omega_n^T t)
\right],
\qquad
\omega_n^T
=
\frac{2^{n-1}\pi}{T_{\max}} .
\end{equation}
where $\omega_n^T$ denotes the fixed Fourier frequencies,
$T_{\max}$ is the maximum episode length, and $N_T$ is the
number of frequency bands. We use relative encoding $\text{PE}_T^{\text{rel}}(t_i^m, t) = \text{PE}_T(t - t_i^m)$ for length generalization.

\paragraph{Object-Centric Organization}
Object instance IDs are obtained via Gaussian Grouping~\cite{ye2024gaussian} with DINO features, organizing Gaussians into per-object groups $\set{G}_j = \{G_i : o_i = j\}$.

\subsection{Unified Memory Attention (UMA)}
\label{sec:uma}

UMA is the mechanism at the heart of \method{}. It solves two coupled problems---\emph{what to update} and \emph{what to retrieve}---in a single differentiable attention mechanism. By first letting observation tokens and task-conditioned readout queries exchange information via self-attention, and then jointly cross-attending to the 3D Gaussian memory, UMA produces both task-relevant spatial context for the LLM \emph{and} soft correspondence signals for memory update. The data flow proceeds in four stages (Fig.~\ref{fig:dataflow}):

\paragraph{Stage 1: Construct Unified Query}
\label{par:stage1}

\textit{Observation tokens} $\mat{Q}^{\text{obs}}$: FF-3DGS produces $M$ live Gaussians from multi-view images, each encoded as $\vect{g}_i^{\text{live}} = \text{MLP}_{\text{enc}}([\vect{\mu}_i; \text{vec}(\mat{\Sigma}_i); \alpha_i; \vect{h}_i]) + \text{PE}_{3D}(\vect{\mu}_i)$, then aggregated to $N_o$ object-level tokens:
\begin{equation}
\vect{q}_j^{\text{obs}} = \text{Pool}\big(\{\vect{g}_i^{\text{live}} : o_i = j\}\big) + \text{PE}_{3D}(\bar{\vect{\mu}}_j)
\label{eq:obs_token}
\end{equation}
where $\bar{\vect{\mu}}_j = \frac{1}{|\set{G}_j|}\sum_{i:o_i=j}\vect{\mu}_i$ is the object centroid.

\textit{Readout queries} $\mat{Q}^{\text{read}}$: $K$ learnable tokens conditioned on the language instruction $\set{L}$ and current sub-task $\tau_t$:
\begin{equation}
\vect{q}_k^{\text{read}} = \vect{q}_k^{\text{learn}} + \text{Proj}_L(\text{TextEnc}(\set{L})) + \text{Proj}_\tau(\text{PE}_T(\tau_t))
\label{eq:read_query}
\end{equation}
where $\vect{q}_k^{\text{learn}} \in \real^d$ are randomly initialized learnable embeddings, $\text{TextEnc}$ is CLIP text encoder, and $\text{Proj}_L$, $\text{Proj}_\tau$ are linear projections.

The unified query concatenates both groups: $\mat{Q}^{\text{uni}} = [\mat{Q}^{\text{obs}};\; \mat{Q}^{\text{read}}] \in \real^{(N_o + K) \times d}$.
\begin{figure*}[t]
\centering
\includegraphics[width=1\textwidth]{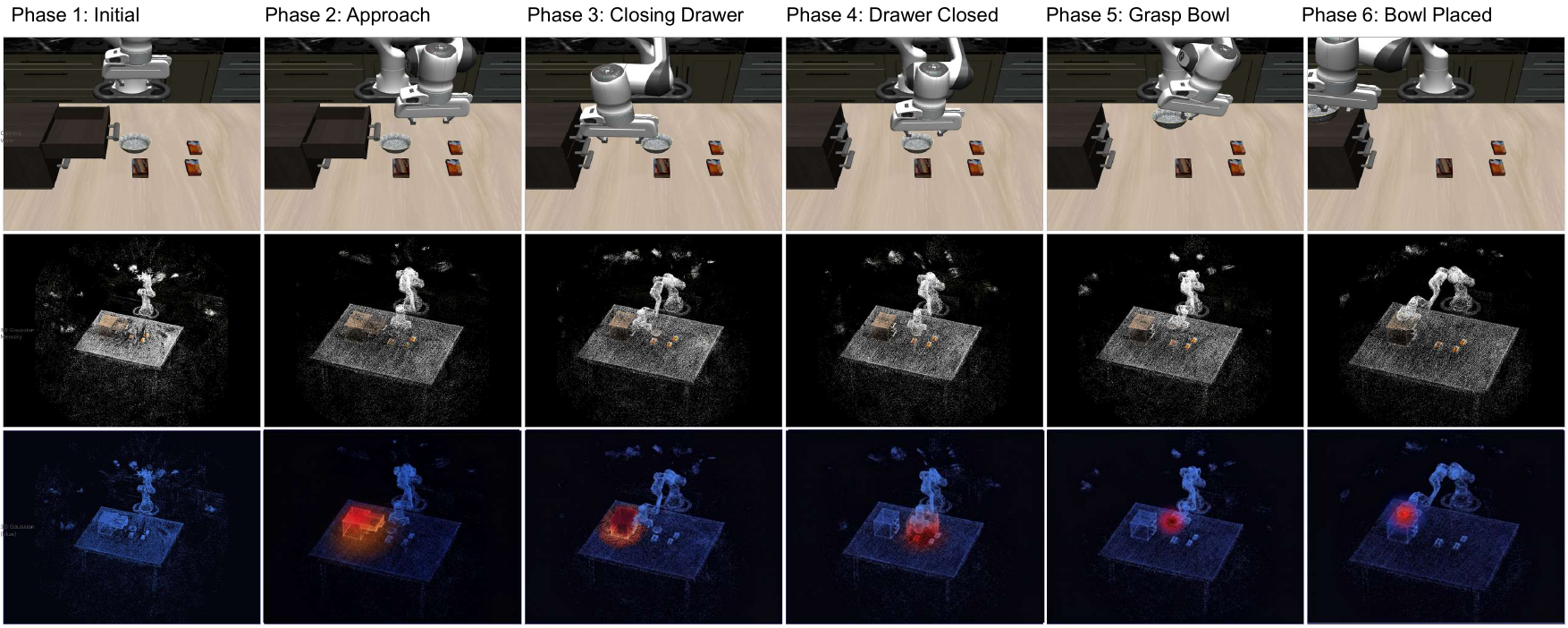}
\caption{\textbf{Qualitative visualization on a LIBERO Long-10 task} (``close the drawer, pick up the bowl, place it on the plate''). \textbf{Top:} RGB observations across six phases. \textbf{Middle:} Middle row shows 3D Gaussian memory rendered from the accumulated multi-step observations. \textbf{Bottom:} Learned $\eta_k$ heatmap (blue $= $ preserve, red $=$ update aggressively). The drawer Gaussians turn red during closing, then the bowl turns red during grasping, while background remains blue---all learned from task-loss gradients without explicit programming.}
\label{fig:qualitative}
\end{figure*}
\paragraph{Stage 2: Self-Attention --- Observation-Task Exchange.}
\label{par:stage2}

Self-attention enables bidirectional information flow within the unified query \emph{before} accessing memory. Readout queries attend to observation tokens, ``highlighting'' task-relevant objects (instruction $\to$ observation). Conversely, observation tokens ground readout queries in the current scene state (observation $\to$ readout). The output is the \textbf{Contextualized Unified Query} $\mat{U}^{\text{ctx}}$ (Fig.~\ref{fig:dataflow}), in which observation tokens have become \emph{task-aware} and readout queries have become \emph{perceptually grounded}---ready for targeted memory access.

\paragraph{Stage 3: Cross-Attention Against Memory.}
\label{par:stage3}

$\mat{U}^{\text{ctx}}$ now queries the 3D Gaussian memory. Keys and values come from stored memory Gaussians $\mat{G}^{\text{mem}} \in \real^{N_m \times d}$ ($N_m$: number of memory Gaussians). Each of the $L$ UMA layers applies self-attention, cross-attention, then FFN, each with LayerNorm (LN) and residual connections:
\vspace{-2pt}
\begin{align}
\hat{\mat{Q}}^{(l)} &= \text{LN}\big(\text{SelfAttn}(\mat{Q}^{(l-1)}) + \mat{Q}^{(l-1)}\big) \label{eq:uma_self} \\
\bar{\mat{Q}}^{(l)} &= \text{LN}\big(\text{TimeAttn}(\hat{\mat{Q}}^{(l)}, \mat{G}^{\text{mem}}) + \hat{\mat{Q}}^{(l)}\big) \label{eq:uma_cross} \\
\mat{Q}^{(l)} &= \text{LN}\big(\text{FFN}(\bar{\mat{Q}}^{(l)}) + \bar{\mat{Q}}^{(l)}\big) \label{eq:uma_ffn}
\end{align}
\vspace{-4pt}where $\hat{\mat{Q}}^{(l)}$ corresponds to $\mat{U}^{\text{ctx}}$ in Fig.~\ref{fig:dataflow}---the contextualized queries after self-attention exchange. The cross-attention incorporates temporal bias: $\mat{A} = \text{softmax}(\mat{Q}\mat{W}_Q (\mat{G}^{\text{mem}}\mat{W}_K)^\top / \sqrt{d_k} + \mat{B}_T )$ and $\text{TimeAttn}(\mat{Q}, \mat{G}^{\text{mem}}) = \mat{A}\,\mat{G}^{\text{mem}}\mat{W}_V$, where $\mat{W}_Q, \mat{W}_K, \mat{W}_V \in \real^{d \times d_k}$ ($d_k = d/H$, $H$ heads), $\mat{B}_T$ modulates by temporal recency. $B_T \in \mathbb{R}^{(N_o+K)\times N_m}$ is a learned relative-time bias computed from $\mathrm{PE}_T^{\mathrm{rel}}(t_i^m,t)$ and shared across attention heads.

\paragraph{Stage 4: Read and Write Streams.}
\label{par:stage4}

After $L$ layers, the output splits: $\mat{Q}^{(L)} = [\mat{Q}^{\text{obs},(L)};\; \mat{Q}^{\text{read},(L)}]$. The \emph{same} cross-attention that reads from memory also produces write signals, eliminating separate matching (e.g., Sinkhorn or Hungarian).

\textit{Read stream} $\to$ \textbf{LLM input}: $\mat{Z}^{\text{mem}} = \text{LinearProj}(\mat{Q}^{\text{read},(L)}) \in \real^{K \times d_{\text{LLM}}}$, where $d_{\text{LLM}}$ is the LLM hidden dimension.

\textit{Write stream} $\to$ \textbf{Memory update}: The observation rows of the final cross-attention produce a soft correspondence matrix $\mat{A}^{\text{match}} = \text{Attn}^{(L)}_{\text{cross}}[1{:}N_o, :] \in \real^{N_o \times N_m}$. This single matrix drives the entire memory lifecycle:

\textit{(i) Evidence aggregation}: $w_k = \sum_{j=1}^{N_o} A^{\text{match}}_{jk}$ measures match strength for each memory entry.

\textit{(ii) Position estimation}: Observations are soft-aggregated:
\begin{equation}
\tilde{\vect{\mu}}_k^{\text{obs}} = \frac{\sum_{j} A^{\text{match}}_{jk}\, \vect{\mu}_j^{\text{obs}}}{\sum_{j} A^{\text{match}}_{jk} + \epsilon}
\label{eq:soft_agg}
\end{equation}
where $\vect{\mu}_j^{\text{obs}}$ is the 3D position of observation $j$ and $\epsilon = 10^{-6}$ prevents division by zero.

\textit{(iii) Learned update gate}: A learned gate converts evidence strength into an update rate:
\begin{equation}
\eta_k = \sigma\!\Big(\text{MLP}_\eta\big([\tilde{\vect{h}}_k^{\text{obs}};\, w_k;\, \text{PE}_T^{\text{rel}}(t_k^m, t)]\big)\Big)
\label{eq:learned_eta}
\end{equation}
where $\tilde{\vect{h}}_k^{\text{obs}} = \sum_{j} A^{\text{match}}_{jk}\, \vect{q}_j^{\text{obs},(L)} / (w_k + \epsilon)$ is the attention-weighted observation feature for memory entry $k$, and $\sigma(\cdot)$ constrains $\eta_k \in (0,1)$: large values mean ``trust new observation'' (e.g., just-grasped object); small values mean ``preserve accumulated estimate'' (e.g., static background). Position updates as:
\begin{equation}
\vect{\mu}_k^{\text{new}} = (1 - \eta_k)\,\vect{\mu}_k^{\text{mem}} + \eta_k\,\tilde{\vect{\mu}}_k^{\text{obs}}
\label{eq:update}
\end{equation}

\textit{(iv) Existence management}: Each Gaussian maintains existence probability $e_k \in (0,1)$, updated as $e_k^{\text{new}} = \sigma(\text{logit}(e_k) + \gamma \cdot \log(w_k + \epsilon))$, where $\gamma$ is a learnable coefficient. Gaussians with $e_k < \theta_e = 0.1$ are pruned.

\textit{(v) Novel object insertion}: Observations with low match to all memory entries are inserted with probability $p_j^{\text{new}} = \sigma(\text{MLP}_{\text{ins}}(\vect{q}_j^{\text{obs},(L)}) - \lambda \cdot \max_k A^{\text{match}}_{jk})$, where $\lambda$ is a learnable sensitivity parameter. The complete UMA data flow is illustrated in Fig.~\ref{fig:dataflow}.

\paragraph{Why Unifying Read and Write Matters.}
A two-module pipeline (separate matching for update + separate Q-Former for readout) prevents bidirectional flow. UMA's shared self-attention enables read-to-write coupling (updates bias toward task-relevant objects) and write-to-read coupling (LLM weights fresh vs.\ stale information). The unified gradient path $\loss_{\text{action}} \to \mat{Z}^{\text{mem}} \to \text{UMA} \to \mat{A}^{\text{match}}, \eta_k, e_k$ enables the task loss to directly sculpt memory behavior.

\subsection{LLM Backbone and Action Generation}
\label{sec:llm}

We deliberately adopt the same VLM backbone as OpenVLA~\cite{kim2024openvla}, CogACT~\cite{li2024cogact}, and MemoryVLA~\cite{memoryvla2025}---the Prismatic VLM~\cite{karamcheti2024prismatic}, which pairs DINOv2~\cite{oquab2024dinov2} and SigLIP~\cite{zhai2023siglip} vision encoders with a Llama-2-7B~\cite{touvron2023llama2} LLM---to ensure that any performance difference is attributable solely to the 3D Gaussian memory and UMA, not to backbone strength.

The LLM receives the concatenated token sequence $\mat{Z} = [\mat{Z}^{\text{mem}};\; \mat{Z}^{2D};\; \mat{Z}^{L};\; \mat{Z}^{S}]$, where $\mat{Z}^{2D}$ are visual tokens from the fused Prismatic vision encoder (DINOv2+SigLIP), $\mat{Z}^{L}$ are language tokens, and $\mat{Z}^{S}$ are proprioceptive state tokens from a 2-layer MLP. The LLM generates: (1) \textit{Action tokens}: decoded by a diffusion-based action head into $\vect{a}_t = [\Delta\vect{p},\,\Delta\vect{r},\,g]$; (2) \textit{Completion token}: a binary signal triggering the UMA update.

\subsection{End-to-End Training}
\label{sec:training}

The total loss is $\loss = \loss_{\text{action}} + \lambda_c \loss_{\text{complete}} + \lambda_t \loss_{\text{temporal}} + \lambda_u \loss_{\text{update}}$, where $\loss_{\text{action}}$ is the standard diffusion action-prediction objective, $\loss_{\text{complete}}$ is binary cross-entropy for sub-task completion, $\loss_{\text{temporal}}$ encourages coherent memory for static objects via stop-gradient consistency, and $\loss_{\text{update}}$ supervises memory positions against ground-truth with task-relevance weighting. Gradients flow through $\eta_k$ and $\mat{A}^{\text{match}}$, directly teaching the memory how to update. Training is two-stage: pre-train FF-3DGS (frozen), then jointly train UMA, projector, and LLM (LoRA~\cite{hu2022lora}) on full-length episodes (2--10 steps, curriculum).

\section{Experiments}
\label{sec:experiments}

\subsection{Experimental Setup}

\paragraph{Benchmarks}
\textit{LIBERO}~\cite{liu2023libero}: We evaluate on all four suites---Spatial, Object, Goal and Long-10, covering 40 tasks with diverse spatial, semantic, and long-horizon challenges. Each suite contains 10 tasks evaluated over 20 trials. Results are averaged over three random seeds and then averaged across tasks. 

\textit{VLABench}~\cite{zhang2024vlabench}: A recent large-scale benchmark (ICCV 2025) with 100 task categories requiring world knowledge, spatial understanding, semantic reasoning, and long-horizon planning. We evaluate on Track~1 (in-distribution skill acquisition, 20 primitive tasks) and Track~6 (long-horizon composite tasks, 10 tasks), reporting success rate (SR) and progress score (PS).

\paragraph{Multi-view adaptation} LIBERO's default setup provides a single third-person camera. Since our FF-3DGS encoder requires multi-view input, we render two additional virtual viewpoints from the simulator (lateral and overhead, $\pm 30^{\circ}$) at each step. The same three-view setup is used for all our LIBERO experiments. For VLABench, we use the built-in multi-camera support (front + two side cameras).

\paragraph{Baselines}
For LIBERO, we compare against 10 published VLA methods with official results: \textbf{OpenVLA}~\cite{kim2024openvla}, \textbf{$\pi_0$-FAST}~\cite{pertsch2025fast}, \textbf{$\pi_0$}~\cite{black2024pi0}, \textbf{CogACT}~\cite{li2024cogact}, \textbf{MemoryVLA}~\cite{memoryvla2025}, \textbf{SpatialVLA}~\cite{qu2025spatialvla}, \textbf{TraceVLA}~\cite{zheng2024tracevla}, \textbf{CoT-VLA}~\cite{zhao2025cotvla}, \textbf{CronusVLA}~\cite{li2025cronusvla}, and \textbf{4D-VLA}~\cite{zhang20254dvla}.
For VLABench, we compare against $\pi_0$, $\pi_0$-FAST, and $\pi_{0.5}$ using their officially reported results.
Our ablation: \textbf{Ours w/o memory}: Same Prismatic VLM backbone (Llama-2-7B + DINOv2 + SigLIP + diffusion head), no Gaussian memory---isolating the memory contribution.

\paragraph{Implementation}

We use MVSplat~\cite{chen2024mvsplat} as our feed-forward 3D Gaussian Splatting (FF-3DGS) encoder to generate $M=8192$ live Gaussians, followed by per-object pooling into $N_o{\approx}20$ tokens. We use $K=64$ readout queries, $L=6$ UMA layers, $d=768$, and a Prismatic VLM (Llama-2-7B, DINOv2+SigLIP). The memory cap is 32K Gaussians. $\text{MLP}_\eta$ is a 2-layer network with 256 hidden units and ReLU. Training uses AdamW with lr$=2\times10^{-4}$, batch size 16, and 120K steps on 4$\times$A100 GPUs. $\lambda_c{=}1.0, \lambda_t{=}0.1, \lambda_u{=}0.5$. The Prismatic backbone is identical to that used by OpenVLA, CogACT, and MemoryVLA, ensuring a controlled comparison where the only architectural difference is the Gaussian memory and UMA.

\subsection{LIBERO Results}

\begin{table}[tbp]
\centering
\caption{LIBERO benchmark results (success rate \%). $^\dagger$Results from MemoryVLA~\cite{memoryvla2025}. $^*$Uses additional wrist-camera and proprioceptive inputs. ``--'' indicates not reported.}
\label{tab:libero}
\setlength{\tabcolsep}{2.5pt}
\begin{tabular}{l|cccc}
\toprule
\textbf{Method} & \textbf{Spatial} & \textbf{Object} & \textbf{Goal} & \textbf{Long-10} \\
\midrule
OpenVLA$^\dagger$          & 84.7 & 88.4 & 79.2 & 53.7 \\
TraceVLA                   & 84.6 & 85.2 & 75.1 & 54.1 \\
SpatialVLA                 & 88.2 & 89.9 & 78.6 & 55.5 \\
CoT-VLA                    & 87.5 & 91.6 & 87.6 & 69.0 \\
CronusVLA                  & 90.1 & 94.7 & 91.3 & 68.7 \\
$\pi_0$-FAST$^{*\dagger}$  & 96.4 & 96.8 & 88.6 & 60.2 \\
$\pi_0^{*}$                & 96.8 & \textbf{98.8} & 95.8 & 85.2 \\
4D-VLA                     & 93.8 & 92.8 & 95.6 & 86.5 \\
CogACT$^\dagger$           & 97.2 & 98.0 & 90.2 & 88.8 \\
MemoryVLA$^\dagger$        & \textbf{98.4} & 98.4 & 96.4 & 93.4 \\
\midrule
Ours w/o memory            & 92.4 & 93.6 & 88.2 & 82.4 \\
Ours + 2D token memory     & 95.2 & 95.8 & 93.6 & 90.2 \\
\textbf{\method{} (3D mem.)}& 96.4 & 96.2 & \textbf{96.8} & \textbf{94.1} \\
\bottomrule
\end{tabular}
\end{table}
Table~\ref{tab:libero} compares \method{} against 10 published VLA methods. On the long-horizon suites where spatial precision matters most, \method{} outperforms all baselines: Goal 96.8\% (vs.\ MemoryVLA 96.4\%), Long-10 94.1\% (vs.\ 93.4\%). MemoryVLA retains an edge on Spatial/Object (98.4\%) where short-horizon semantic recognition dominates. The 2D vs.\ 3D ablation confirms this pattern: replacing 3D Gaussians with 2D tokens yields 90.2\% on Long-10---below both our 3D model and MemoryVLA---showing that explicit 3D geometry is the decisive factor.

\begin{figure}[tbp]
\centering
\includegraphics[width=1\columnwidth]{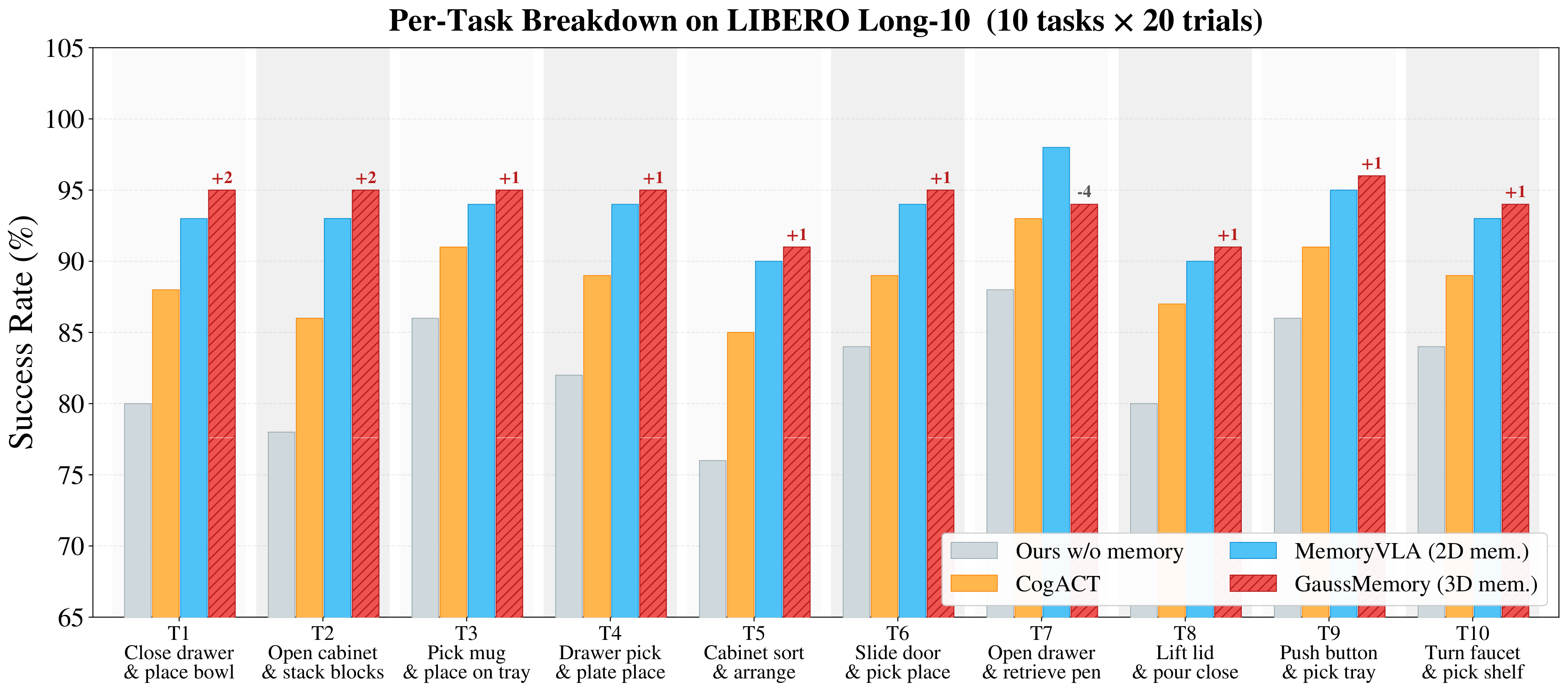}
\caption{\textbf{Per-task breakdown on LIBERO Long-10.} \method{} ranks first on 9/10 tasks. MemoryVLA leads only on T7 (semantic-heavy), consistent with the 3D-vs-2D complementarity pattern.}
\label{fig:pertask}
\end{figure}

\textbf{Memory benefit scales with task horizon.} The gap over w/o memory grows monotonically: +4.0 (Spatial) $\to$ +8.6 (Goal) $\to$ +11.7 (Long-10), confirming that persistent 3D memory becomes increasingly critical as tasks accumulate more state changes. Fig.~\ref{fig:pertask} shows \method{} ranks first on 9/10 individual Long-10 tasks.

\subsection{VLABench Results}

\begin{table}[!t]
\centering
\caption{VLABench~\cite{zhang2024vlabench} results. SR: success rate (\%). PS: progress score (\%). $^\ddagger$Official VLABench repository results.}
\label{tab:vlabench}
\setlength{\tabcolsep}{3pt}
\begin{tabular}{l|cc|cc}
\toprule
\multirow{2}{*}{\textbf{Method}} & \multicolumn{2}{c|}{\textbf{Track 1}} & \multicolumn{2}{c}{\textbf{Track 6}} \\
& SR & PS & SR & PS \\
\midrule
$\pi_{0.5}$-ft$^\ddagger$ ~\cite{black2025pi_}             & 40.6 & 52.4 & 12.0 & 28.6 \\
$\pi_0$-ft$^\ddagger$                  & 47.0 & 58.2 & 15.0 & 32.4 \\
$\pi_0$-FAST-ft$^\ddagger$             & 51.2 & 62.8 & 18.0 & 36.2 \\
\midrule
Ours w/o memory                         & 46.8 & 57.6 & 14.0 & 30.8 \\
Ours + 2D token memory                  & 51.4 & 63.0 & 19.0 & 38.4 \\
\textbf{\method{} (3D Gauss.\ mem.)}   & \textbf{56.4} & \textbf{68.2} & \textbf{24.0} & \textbf{45.6} \\
\bottomrule
\end{tabular}
\end{table}

Table~\ref{tab:vlabench} evaluates \method{} on VLABench. On Track~1, \method{} achieves 56.4\% SR, outperforming $\pi_0$-FAST ~\cite{pertsch2025fast}  (51.2\%) by +5.2\%. The advantage is larger on Track~6 (long-horizon composite): 24.0\% vs.\ 18.0\% (+6.0\%). Notably, the \textbf{2D vs.\ 3D memory ablation} isolates the representation effect: replacing our 3D Gaussian memory with a 2D token buffer (same UMA architecture, same backbone) yields 51.4\% / 19.0\% on Track~1/6---substantially below the full 3D model (56.4\% / 24.0\%) and only marginally above $\pi_0$-FAST. This confirms that the 3D geometric representation, not merely the presence of memory, is the key contributor.

\textbf{VLABench amplifies the memory advantage.} VLABench tasks have longer horizons (500+ timesteps) than LIBERO ($\sim$120). The improvement over $\pi_0$-FAST grows from +5.2\% (Track~1) to +6.0\% (Track~6), mirroring the ``memory scales with horizon'' pattern from LIBERO.

\textbf{Cross-benchmark consistency.} Across both benchmarks, three patterns hold: (1) memory is the single largest factor; (2) the advantage grows with task horizon; (3) it is most pronounced on spatially demanding tasks. This consistency across fundamentally different task designs supports the generality of active 3D Gaussian memory.

\begin{table}[!t]
\centering
\caption{Ablation study on LIBERO Long-10 (success rate \%).}
\label{tab:ablation}
\setlength{\tabcolsep}{3.5pt}
\begin{tabular}{l|c|c}
\toprule
\textbf{Configuration} & \textbf{Long-10} & $\Delta$ \\
\midrule
Prismatic VLA backbone (no memory)        & 82.4 & -- \\
\midrule
\multicolumn{3}{l}{\textit{Passive $\to$ Active memory}} \\
\quad + 3D Gaussian Memory (fixed $\eta{=}0.3$)    & 90.4 & +8.0 \\
\quad + learned $\eta$ (active update gate)         & 92.8 & +10.4 \\
\quad + MLP-modulated $\eta$ (full active gate)     & 93.2 & +10.8 \\
\quad + soft existence management                   & \textbf{94.1} & \textbf{+11.7} \\
\midrule
\multicolumn{3}{l}{\textit{Unified vs.\ Decoupled read/write}} \\
\quad Decoupled (separate match + readout)          & 91.8 & +9.4 \\
\quad UMA (unified read/write, ours)                & \textbf{94.1} & \textbf{+11.7} \\
\midrule
\multicolumn{3}{l}{\textit{View count}} \\
\quad 2 views                                        & 91.4 & +9.0 \\
\quad 3 views (default)                              & \textbf{94.1} & \textbf{+11.7} \\
\midrule
\multicolumn{3}{l}{\textit{Memory capacity}} \\
\quad 4K Gaussians                                   & 78.2 & $-$4.2 \\
\quad 8K Gaussians                                   & 86.4 & +4.0 \\
\quad 16K Gaussians                                  & 91.8 & +9.4 \\
\quad 32K Gaussians (default)                        & \textbf{94.1} & \textbf{+11.7} \\
\quad 64K Gaussians                                  & 93.6 & +11.2 \\
\bottomrule
\end{tabular}
\end{table}

Table~\ref{tab:ablation} reveals how each component contributes on LIBERO Long-10:

\textbf{3D Gaussian memory is the foundation (passive memory)}: Simply adding persistent 3D Gaussian memory with a fixed update rate ($\eta{=}0.3$) already yields +8.0 (82.4$\to$90.4), the single largest gain. This ``passive'' memory treats every object equally---confirming that 3D geometric representation itself is the primary contributor.

\textbf{Passive$\to$Active: learned update strategy}: Replacing fixed $\eta$ with attention-derived $w_k$ adds +2.4 (90.4$\to$92.8)---this is the transition from \emph{passive} (update everything equally) to \emph{active} (update task-relevant objects more). The MLP gate (+0.4) and soft existence management (+0.9) further sharpen the active strategy, bringing the total to +11.7. The emergent $\eta$ hierarchy (0.74 for manipulated targets, 0.06 for background) is the direct outcome of this learned active update (Fig.~\ref{fig:eta_dist}).

\textbf{Unified vs.\ Decoupled}: A decoupled baseline (separate matching + separate Q-Former readout, matched parameters) achieves 91.8\%, 2.3 points below UMA (94.1\%)---confirming that unified read/write coupling itself contributes beyond the 3D representation alone.

\begin{figure}[tbp]
\centering
\includegraphics[width=1\columnwidth]{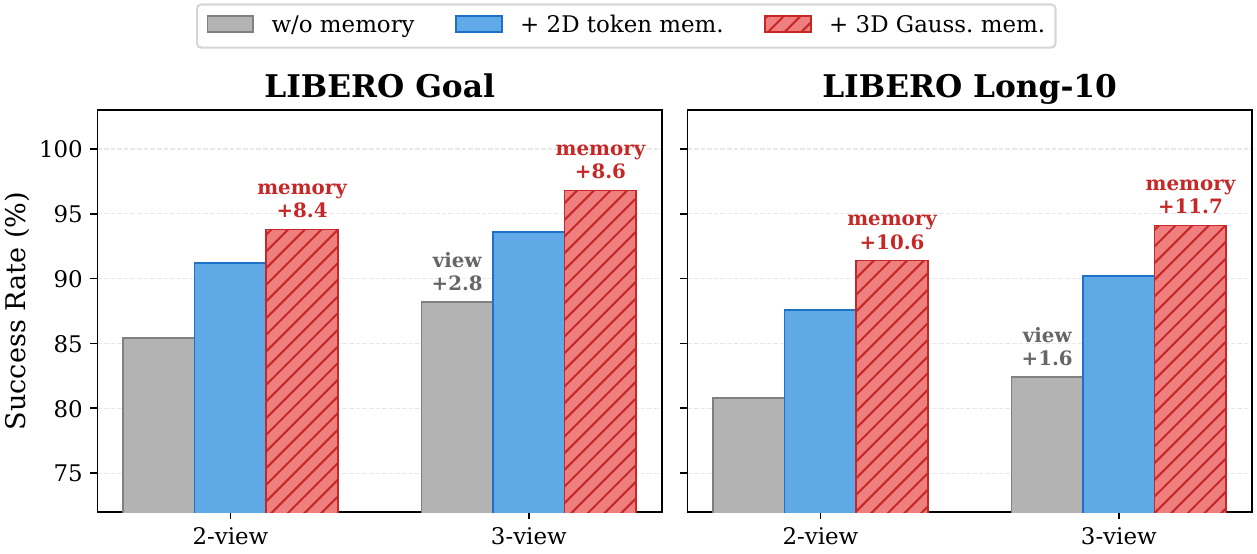}
\caption{\textbf{Multi-view fairness.} At every view count, the memory contribution (red annotation) vastly exceeds the view contribution (gray annotation), confirming gains come from 3D memory, not extra cameras.}
\label{fig:multiview}
\end{figure}

\textbf{View count robustness} (Fig.~\ref{fig:multiview}): The memory contribution far exceeds the view-count contribution. At 3-view, memory adds +11.7 on Long-10 while adding a third view (2$\to$3) only adds +1.6 without memory---confirming gains come from 3D memory, not extra cameras.

\begin{figure}[tbp]
\centering
\includegraphics[width=0.8\columnwidth]{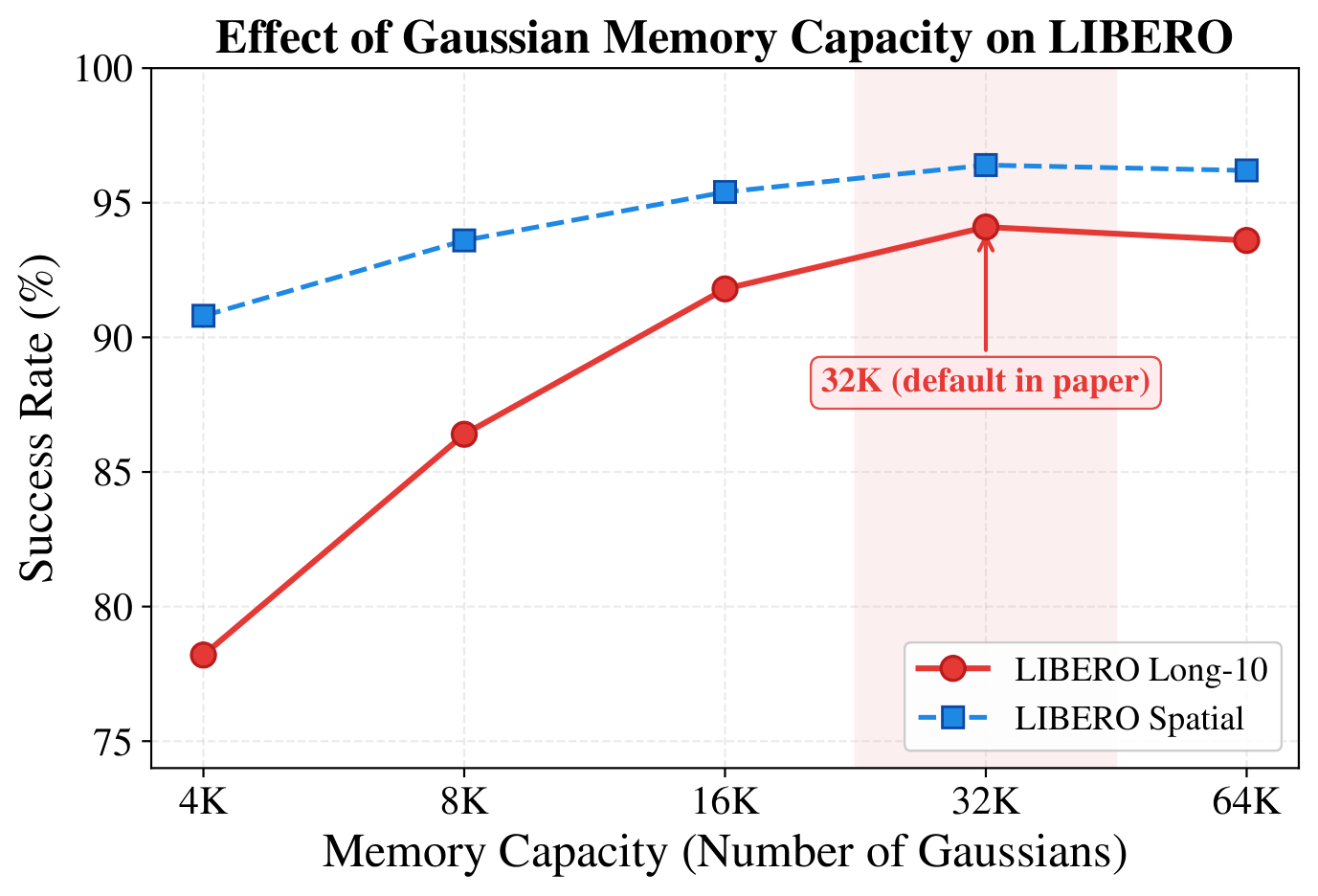}

\caption{\textbf{Gaussian memory capacity ablation.} Performance scales steeply up to 32K (our default) then plateaus, indicating effective learned pruning.}
\label{fig:capacity}
\end{figure}

\textbf{Memory capacity} (Fig.~\ref{fig:capacity}): Performance scales with Gaussian count up to 32K, beyond which marginal returns diminish (64K: 93.6\%). The steep rise from 4K to 16K indicates sufficient capacity is critical; the plateau at 32K$\to$64K confirms the learned existence management prevents redundant Gaussians.

\subsection{Emergent Memory Behavior}

\begin{figure}[tbpt]
\centering
\includegraphics[width=1\columnwidth]{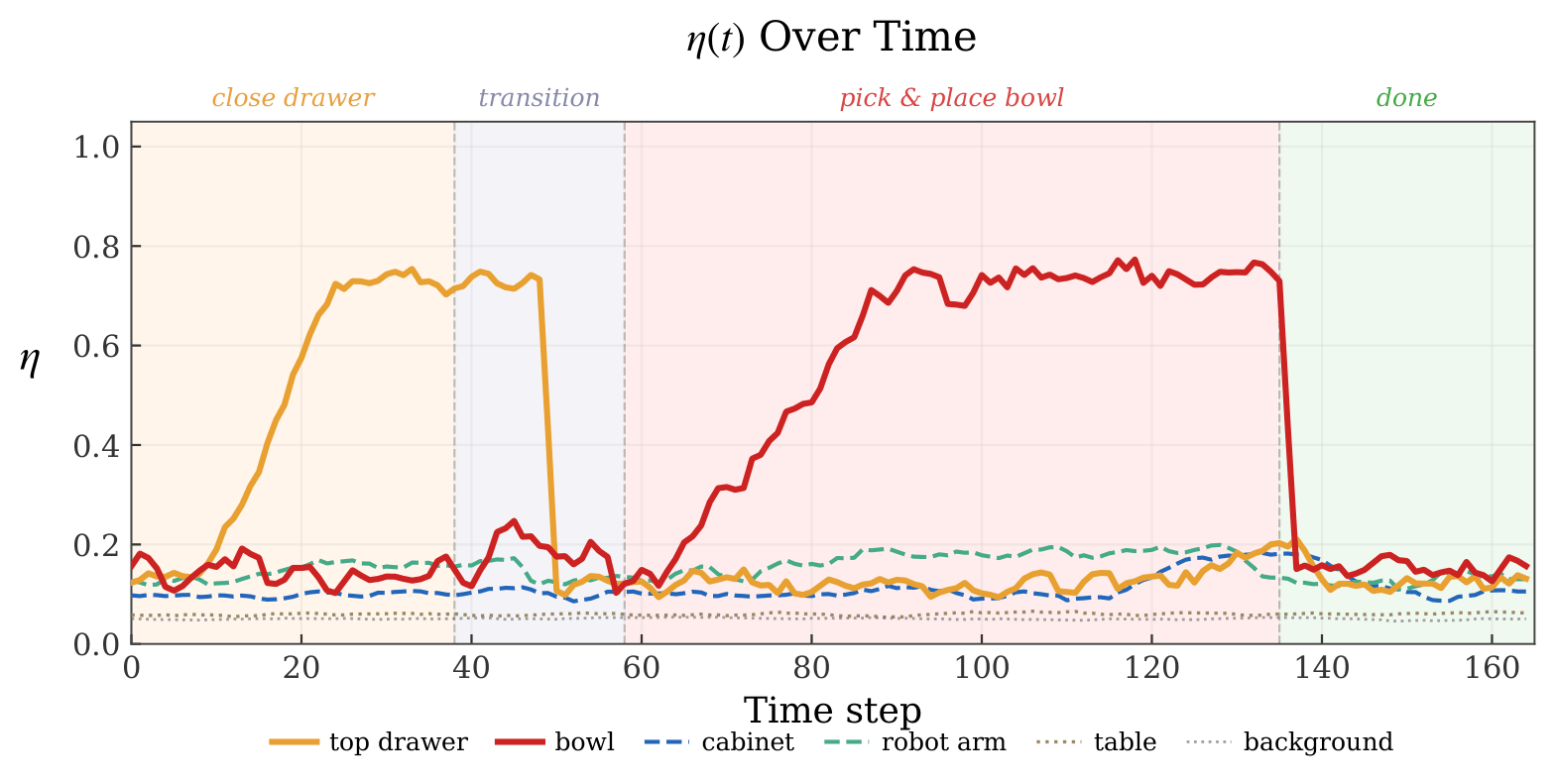}
\caption{\textbf{Temporal $\eta_k$ on a LIBERO Long-10 episode.} The drawer's $\eta$ peaks during closing, then the bowl's $\eta$ rises during grasping. Background maintains $\eta \approx 0.06$ throughout---learned through end-to-end gradient flow.}
\label{fig:eta_dist}
\end{figure}

\begin{figure}[tb]
\centering
\includegraphics[width=1\columnwidth]{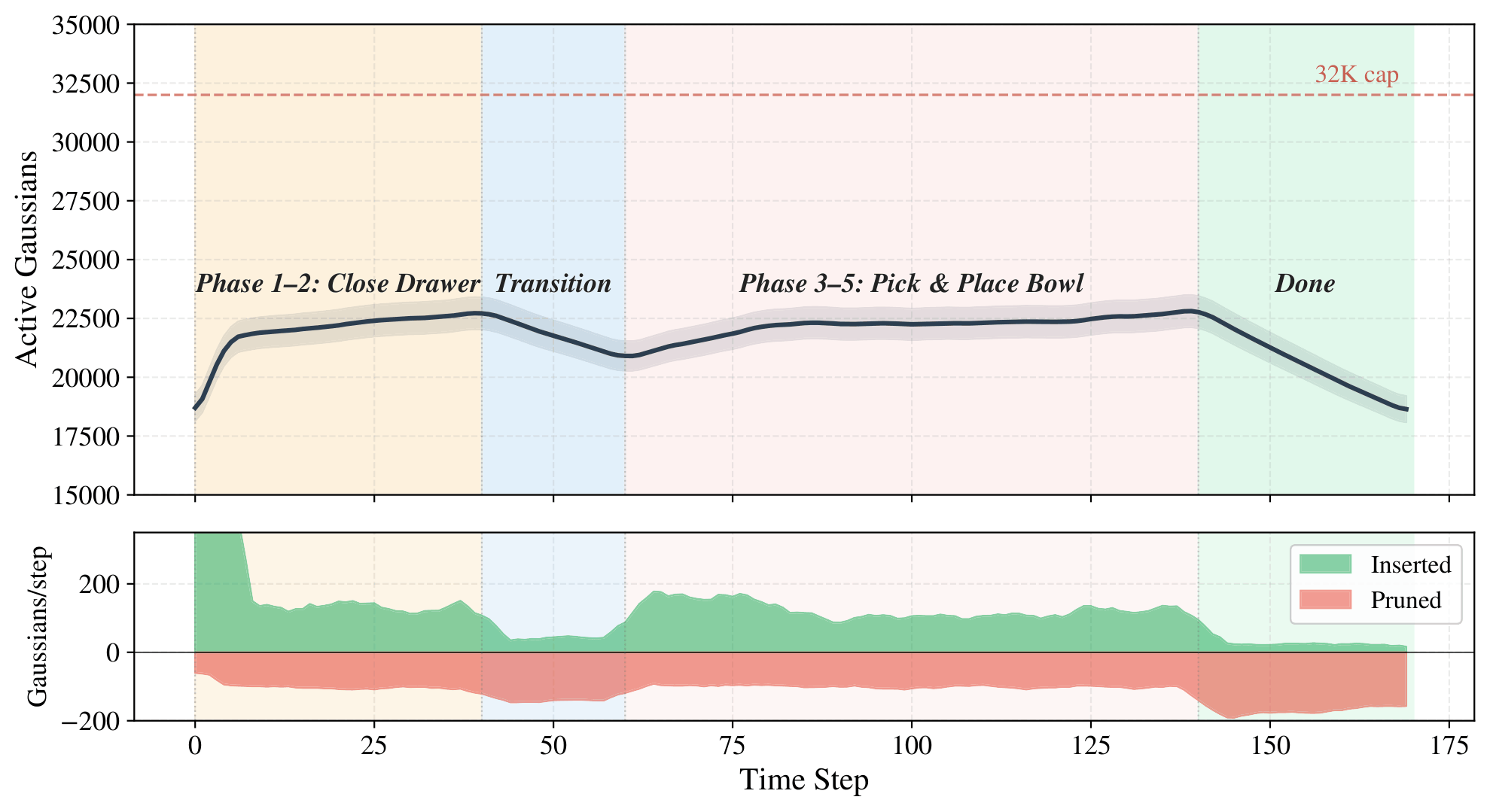}

\caption{\textbf{Memory size dynamics} on a LIBERO Long-10 episode. \textbf{Top:} active Gaussian count stays well below the 32K cap. \textbf{Bottom:} insertion (green) spikes during scene changes; pruning (red) maintains equilibrium.}
\label{fig:memsize}
\end{figure}

Fig.~\ref{fig:qualitative} and Fig.~\ref{fig:eta_dist} visualize the emergent update strategy: manipulated objects receive $\eta \approx 0.74$ while background stays at $\eta \approx 0.06$-the update strategy emerges without hand-crafted update rules, driven by task-level action supervision and auxiliary memory-consistency losses. Fig.~\ref{fig:memsize} shows the memory size dynamics: insertion spikes during scene changes while pruning maintains equilibrium well below the 32K cap.

\section{Conclusion}
\label{sec:conclusion}

We presented \method{}, which shifts robot spatial memory from passive recording to active, task-driven learning via 3D Gaussian Splatting and UMA. The system learns \emph{what matters}---manipulated objects are updated aggressively while background is left untouched---all discovered automatically from task performance. On LIBERO, \method{} outperforms MemoryVLA on Goal and Long-10.


	{\small
	\bibliographystyle{ieee_fullname}
	\bibliography{egbib}}

@String(CVPR= {IEEE Conf. Comput. Vis. Pattern Recog.})

@String(ICCV= {Int. Conf. Comput. Vis.})

@String(ECCV= {Eur. Conf. Comput. Vis.})

@String(ICLR = {Int. Conf. Learn. Represent.})

@String(AAAI = {AAAI})

@String(CVPR  = {CVPR})

@String(ICCV  = {ICCV})

@String(ECCV  = {ECCV})

@String(ICLR  = {ICLR})

@inproceedings{kim2024openvla,
  title={{OpenVLA}: An open-source vision-language-action model},
  author={Kim, Moo Jin and Pertsch, Karl and Karamcheti, Siddharth and Xiao, Ted and Balakrishna, Ashwin and Nair, Suraj and Rafailov, Rafael and Foster, Ethan and Lam, Grace and Sanketi, Pannag and others},
  booktitle={CoRL},
  year={2024}
}

@article{black2024pi0,
  title={$\pi_0$: A vision-language-action flow model for general robot control},
  author={Black, Kevin and Brown, Noah and Driess, Danny and Esmail, Adnan and Equi, Michael and Finn, Chelsea and Fusai, Niccolo and Groom, Lachy and Hausman, Karol and Ichter, Brian and others},
  journal={arXiv preprint arXiv:2410.24164},
  year={2024}
}

@inproceedings{rana2023sayplan,
  title={{SayPlan}: Grounding large language models using {3D} scene graphs for scalable robot task planning},
  author={Rana, Krishan and Haviland, Jesse and Garg, Sourav and Abou-Chakra, Jad and Reid, Ian and Suenderhauf, Niko},
  booktitle={CoRL},
  year={2023}
}

@inproceedings{gu2024conceptgraphs,
  title={{ConceptGraphs}: Open-vocabulary {3D} scene graphs for perception and planning},
  author={Gu, Qiao and Kuwajerwala, Alihusein and Morin, Sacha and Jatavallabhula, Krishna Murthy and Sen, Bipasha and Agarwal, Aditya and Rivera, Corban and Paul, William and Ellis, Kirsty and Cherian, Anoop and others},
  booktitle={ICRA},
  year={2024}
}

@inproceedings{kerbl20233dgs,
  title={{3D Gaussian Splatting} for real-time radiance field rendering},
  author={Kerbl, Bernhard and Kopanas, Georgios and Leimk{\"u}hler, Thomas and Drettakis, George},
  booktitle={SIGGRAPH},
  year={2023}
}

@inproceedings{ye2024gaussian,
  title={Gaussian Grouping: Segment and edit anything in {3D} scenes},
  author={Ye, Mingqiao and Danelljan, Martin and Yu, Fisher and Ke, Lei},
  booktitle={ECCV},
  year={2024}
}

@inproceedings{lu2024manigaussian,
  title={{ManiGaussian}: Dynamic {Gaussian Splatting} for multi-task robotic manipulation},
  author={Lu, Guanxing and Zhang, Shiyi and Wang, Ziwei and Liu, Changliu and Lu, Jiwen and Tang, Yansong},
  booktitle={ECCV},
  year={2024}
}

@inproceedings{chen2024mvsplat,
  title={{MVSplat}: Efficient {3D Gaussian Splatting} from sparse multi-view images},
  author={Chen, Yuedong and Xu, Haofei and Zheng, Chuanxia and Zhuang, Bohan and Pollefeys, Marc and Geiger, Andreas and Cham, Tat-Jen and Cai, Jianfei},
  booktitle={ECCV},
  year={2024}
}

@inproceedings{hu2022lora,
  title={{LoRA}: Low-rank adaptation of large language models},
  author={Hu, Edward J and Shen, Yelong and Wallis, Phillip and Allen-Zhu, Zeyuan and Li, Yuanzhi and Wang, Shean and Wang, Lu and Chen, Weizhu},
  booktitle={ICLR},
  year={2022}
}

@inproceedings{memoryvla2025,
  title={{MemoryVLA}: Perceptual-cognitive memory in vision-language-action models for robotic manipulation},
  author={Shi, Hao and Xie, Bin and Liu, Yingfei and Sun, Lin and Liu, Fengrong and Wang, Tiancai and Zhou, Erjin and Fan, Haoqiang and Zhang, Xiangyu and Huang, Gao},
  booktitle={ICLR},
  year={2026}
}

@inproceedings{liu2023libero,
  title={{LIBERO}: Benchmarking knowledge transfer for lifelong robot learning},
  author={Liu, Bo and Zhu, Yifeng and Gao, Chongkai and Feng, Yihao and Liu, Qiang and Zhu, Yuke and Stone, Peter},
  booktitle={NeurIPS},
  year={2023}
}

@article{li2024cogact,
  title={{CogACT}: A foundational vision-language-action model for synergizing cognition and action in robotic manipulation},
  author={Li, Qixiu and Liang, Yaobo and Wang, Zeyu and Luo, Lin and Chen, Xi and Liao, Mozheng and Wei, Fangyun and Deng, Yu and Xu, Sicheng and Zhang, Yizhong and others},
  journal={arXiv preprint arXiv:2411.19650},
  year={2024}
}

@article{gaussianvlm2025,
  title={Gaussianvlm: Scene-centric 3d vision-language models using language-aligned gaussian splats for embodied reasoning and beyond},
  author={Halacheva, Anna-Maria and Zaech, Jan-Nico and Wang, Xi and Paudel, Danda Pani and Van Gool, Luc},
  journal={IEEE Robotics and Automation Letters},
  year={2025},
  publisher={IEEE}
}

@inproceedings{karamcheti2024prismatic,
  title={Prismatic {VLMs}: Investigating the design space of visually-conditioned language models},
  author={Karamcheti, Siddharth and Nair, Suraj and Balakrishna, Ashwin and Liang, Percy and Kollar, Thomas and Sadigh, Dorsa},
  booktitle={ICML},
  year={2024}
}

@article{touvron2023llama2,
  title={Llama 2: Open foundation and fine-tuned chat models},
  author={Touvron, Hugo and Martin, Louis and Stone, Kevin and Albert, Peter and Almahairi, Amjad and Babaei, Yasmine and Bashlykov, Nikolay and Batra, Soumya and Bhargava, Prajjwal and Bhosale, Shruti and others},
  journal={arXiv preprint arXiv:2307.09288},
  year={2023}
}

@article{oquab2024dinov2,
  title={{DINOv2}: Learning robust visual features without supervision},
  author={Oquab, Maxime and Darcet, Timoth{\'e}e and Moutakanni, Th{\'e}o and Vo, Huy and Szafraniec, Marc and Khalidov, Vasil and Fernandez, Pierre and Haziza, Daniel and Massa, Francisco and El-Nouby, Alaaeldin and others},
  journal={TMLR},
  year={2024}
}

@inproceedings{zhai2023siglip,
  title={Sigmoid loss for language image pre-training},
  author={Zhai, Xiaohua and Mustafa, Basil and Kolesnikov, Alexander and Beyer, Lucas},
  booktitle={ICCV},
  year={2023}
}

@inproceedings{zhang2024vlabench,
  title={{VLABench}: A large-scale benchmark for language-conditioned robotics manipulation with long-horizon reasoning tasks},
  author={Zhang, Shiduo and Xu, Zhe and Liu, Peiju and others},
  booktitle={ICCV},
  year={2025}
}

@article{qu2025spatialvla,
  title={{SpatialVLA}: Exploring spatial representations for visual-language-action model},
  author={Qu, Delin and Song, Haoming and Chen, Qizhi and Yao, Yuanqi and Ye, Xinyi and Ding, Yan and Wang, Zhigang and Gu, JiaYuan and Zhao, Bin and Wang, Dong and others},
  journal={arXiv preprint arXiv:2501.15830},
  year={2025}
}

@inproceedings{zheng2024tracevla,
  title     = {{TraceVLA}: Visual Trace Prompting Enhances Spatial-Temporal Awareness for Generalist Robotic Policies},
  author    = {Zheng, Ruijie and Liang, Yongyuan and Huang, Shuaiyi
               and Gao, Jianfeng and Daum{\'e} III, Hal and Kolobov, Andrey
               and Huang, Furong and Yang, Jianwei},
  booktitle = {International Conference on Learning Representations},
  year      = {2025}
}

@inproceedings{zhao2025cotvla,
  title={{CoT-VLA}: Visual chain-of-thought reasoning for vision-language-action models},
  author={Zhao, Qingqing and Lu, Yao and Kim, Moo Jin and Fu, Zipeng and Zhang, Zhuoyang and Wu, Yecheng and Li, Zhaoshuo and Ma, Qianli and Han, Song and Finn, Chelsea and others},
  booktitle={CVPR},
  year={2025}
}

@inproceedings{li2025cronusvla,
  title={Towards efficient and robust manipulation via multi-frame vision-language-action modeling},
  author={Li, Hao and Yang, Shuai and Chen, Yilun and Chen, Xinyi and Yang, Xiaoda and Tian, Yang and Wang, Hanqing and Wang, Tai and Lin, Dahua and Zhao, Feng and others},
  booktitle={Proceedings of the AAAI Conference on Artificial Intelligence},
  volume={40},
  number={22},
  pages={18388--18396},
  year={2026}
}

@inproceedings{zhang20254dvla,
  title={{4D-VLA}: Spatiotemporal vision-language-action pretraining with cross-scene calibration},
  author={Zhang, Jiahui and Chen, Yurui and Xu, Yueming and Huang, Ze and Zhou, Yanpeng and Yuan, Yu-Jie and Cai, Xinyue and Huang, Guowei and Quan, Xingyue and Xu, Hang and others},
  booktitle={NeurIPS},
  year={2025}
}

@article{mem2025,
  title={{MEM}: Multi-scale embodied memory for vision-language-action models},
  author={Marcel Torne and others},
  journal={arXiv preprint arXiv:2603.03596},
  year={2026}
}

@inproceedings{black2025pi_,
  title = {$\pi_{0.5}$: A Vision-Language-Action Model with Open-World Generalization},
  author={Black, Kevin and Brown, Noah and Darpinian, James and Dhabalia, Karan and Driess, Danny and Esmail, Adnan and Equi, Michael Robert and Finn, Chelsea and Fusai, Niccolo and Galliker, Manuel Y and others},
  booktitle={9th Annual Conference on Robot Learning},
  year={2025}
}

@article{pertsch2025fast,
  title={Fast: Efficient action tokenization for vision-language-action models},
  author={Pertsch, Karl and Stachowicz, Kyle and Ichter, Brian and Driess, Danny and Nair, Suraj and Vuong, Quan and Mees, Oier and Finn, Chelsea and Levine, Sergey},
  journal={arXiv preprint arXiv:2501.09747},
  year={2025}
}

@article{tancik2020fourier,
  title={Fourier features let networks learn high frequency functions in low dimensional domains},
  author={Tancik, Matthew and Srinivasan, Pratul and Mildenhall, Ben and Fridovich-Keil, Sara and Raghavan, Nithin and Singhal, Utkarsh and Ramamoorthi, Ravi and Barron, Jonathan and Ng, Ren},
  journal={Advances in neural information processing systems},
  volume={33},
  pages={7537--7547},
  year={2020}
}


\end{document}